# Keyphrase Based Arabic Summarizer (KPAS)

Tarek El-Shishtawy[1] and Fatma El-Ghannam[2]

[1] Benha University
Cairo, Faculty of Engineering (Shoubra), Egypt
*shishtawy@hotmil.com*

[2] Electronics Research Institute
Cairo, Egypt
*elghannamf@hotmail.com*

**Abstract**
This paper describes a computationally inexpensive and efficient generic summarization algorithm for Arabic texts. The algorithm belongs to extractive summarization family, which reduces the problem into representative sentences identification and extraction sub-problems. Important keyphrases of the document to be summarized are identified employing combinations of statistical and linguistic features. The sentence extraction algorithm exploits keyphrases as the primary attributes to rank a sentence. The present experimental work, demonstrates different techniques for achieving various summarization goals including: informative richness, coverage of both main and auxiliary topics, and keeping redundancy to a minimum. A scoring scheme is then adopted that balances between these summarization goals. To evaluate the resulted Arabic summaries with well-established systems, aligned English/Arabic texts are used through the experiments.

## 1. Introduction

A summary can be defined as a text that is produced from one or more texts that contains a significant portion of the information in the original text [11]. As the number of electronic documents rapidly increases, the need for automatic techniques to assess the relevance of documents materializes. Summaries could be displayed in search engine web pages results as an informative tool for the user to find the relevant and desired information.

Summarization process can be classified according to many directives: method, granularity, generality, and information richness. In general, there are two methods for automatic text summarization: extractive and abstractive. Extractive summarization involves copying significant units (usually sentences) of the original documents. The goal of abstraction summary is to produce summaries that read as text produced by humans. Therefore, abstraction summary may need the building of an internal semantic representation, the use of natural language generation techniques, the compression of sentences, the reformulation, or the use of new word sequences that are not present in the original document. These methods are still difficult to achieve [8]. Extractive summaries can have different granularity levels, which reflect the size of text unit to be extracted, starting from word, phrase distribution, and up to complete paragraph extraction. Concerning generality of summaries, two types can be distinguished: generic summaries and query-driven summaries. The first type tries to represent all relevant topics of a source text. The second type focuses on the user's desired query keywords or topics. Another classification of summaries is based on its information richness. Indicative summaries give a brief idea of what the original text is about without conveying specific content. It is used to indicate important topics to quickly decide whether a text is worth reading. The second type is informative summaries, which are intended to cover the topics in the source text to provide some shortened version of the content. [16].

In this paper we present a computationally inexpensive and efficient generic summarization technique that focuses on keyphrase-based extractive summarization. Keyphrases that are automatically extracted from a document text are used to evaluate the importance of each sentence in the document. Although there are numerous techniques for sentence level extraction, little attention is paid to changing extraction strategy to achieve one or more summarization goals. A human summarizer has the ability to select the sentences to be presented according to many factors including the maximum allowed number of sentences to be displayed and the number of topics to be covered in the document.. Also, a human summmarizer may change the selection strategy if it was noticed that the document contains equal important or main-supplementary concepts. Therefore, the main objective of our work is to demonstrate, using smaller language constructs (keyphrases), a more flexibility method in directing the proposed sentence extractor towards one or more summarization goals. The goals are:

- Extract the most informative sentences that capture main topics. This may be useful in extracting very short summary
- Eliminate the domination of main topic on output summary. This is particularly important for documents that contain main and supplementing topics.
- Keep sentence redundancy to a minimum. This is an essential requirement for a summarizer to allow a room for other concepts to be presented in the output.
- Cover all important topics of the document. For long documents, this puts the lower limit of compression.
- Achieve balance between all previous goals.

The presented keyphrase extraction and summarization heuristics are language-independent, however it is implemented to extract summaries of Arabic documents. To the best of our knowledge, this is the first work that exploits automatically extracted keyphrases to produce Arabic summaries.

The remaining of this paper is organized as follows: the "Related works" is presented in section 2 and a fast revision of the accurate Arabic lemmatizer is presented in section 3. In keyphrase based summarization, the efficiency of the summary is mainly determined by the keyphrase extraction sub-system. Therefore, one of the main contributions of this work is the modification done to an existing Arabic keyphrase extraction subsystem [6] that helps to select the relevant sentences to be in the summary. In section 4, the keyphrase extraction algorithm is reviewed. The paper proposes the use of new heuristics to identify significant sentences. Each heuristic has its pros and cons that affect the summarization results. Section 5 discusses these effects and introduces a combined heuristic to achieve a balance between summarization goals.

## 2. Related works

The basic idea of the extraction is to create summaries using words, phrases, sentences, and even paragraphs pulled directly from the source text using statistical and linguistic analysis.

Word level summary has been started as early as in the 1950's. Luhn [14] introduced a way for summarization based only on word distribution in the source document. This representation abstracts the source text into a frequency table. Similar approaches still used today to generate tag cloud interface [25]. The focus is to present nouns or adjective-noun pairs frequently mentioned in the reviews with font size proportional to their number of occurrences.

Keyphrases are the second representation level in abstraction hierarchy. Keyphrases are defined as a short list of terms to provide condensed summary of the topics of a document [21], where important words and phrases that express the main topics are extracted. D'Avanzo et. al., [22] proposed the LAKE system to extract keyphrases. The summary is presented as list of keyphrases that approximates the summarization.

Most of the extractive summarization techniques consider sentence as a basic extraction unit. Earlier techniques were based on surface level features, such as occurrence of cue phrase markers (eg. "in conclusion", "in summary") [3], Other researchers rely on formatting attributes, such as position of sentences, bold texts, or headers [12].

Another class of summarization depends on scoring sentences, which takes many forms based on statistical, syntactic or semantic attributes. Relevance Measure (RM) and Latent Semantic Analysis (LSA) were used to score sentences [9]. The motivation is to identify topics and evaluate the importance of these topics. Lexical chaining algorithm is another way to group the words of a given text, so that the words of each group has a close semantic relation [15], the motivation is to identify topics and evaluate the importance of these topics. Chains are created by taking a new text word and finding a related chain for it according to relatedness criteria. Barzilay and Elhadad [1] introduced an algorithm to compute lexical chains in a text, merging several knowledge sources such as WordNet thesaurus, a part-of-speech tagger and shallow parser for the identification of nominal groups. The chain representation approach avoids selecting different terms having the same concept (using wordnet) problem, because all these terms occur in the same chain, which reflects that they represent the same concept.

Goldstein [8] presented a summarizer for news-article generated by sentence selection. Sentences are directly ranked for potential inclusion in the summary using a weighted combination of statistical and linguistic features. An English domain-specific sentence-scoring technique was presented in [17]. The system ranks a sentence based on sentence position, keyphrase existence, cue phrase existence, sentence length, and sentence similarity to document class.

Some research has treated sentence extraction as a learning problem [13]. In this approach, training material exemplifies extracted sentences by providing sentence features and selection flag.

The study of Sobh et. al, [18] introduced an Arabic classifier that is trained to extract important sentences directly based on many features such as sentence length, sentence position in paragraph, sentence similarity, number of infinitives in sentence and number of verbs in sentences. An Interactive Document Summarizer (IDS) using automatically extracted keyphrases is introduced by Steve et.al,[20]. Keyphrases are extracted from a document using the Kea keyphrase extraction algorithm [7]. Each sentence in each document is then awarded a score, using a heuristic based on the frequency of the document's keyphrases in the sentence. A further characteristic of an interactive summarizer is the ability to bias the summary towards a particular topic or set of topics.

The methods described above belong to a family of techniques that rely on word, phrase, or sentence distribution, where concept is a single or multiple words. In Salton et al. [19], paragraphs are represented as vectors then the similarity between paragraphs is evaluated. A graph of paragraphs as nodes and the similarities as links is constructed. Given a threshold for similarity the link between two paragraphs exists only if the similarity is above that threshold. The summarization is based on paragraph selection heuristics that Selects paragraphs with many connections with other paragraphs and present them in text order.

## 3. The Proposed System

In this paper, we propose an algorithm that belongs to extractive summarization family, which reduces the problem into a sentence scoring and extraction sub-problems. Instead of scoring sentences directly, keyphrases are used as attributes to evaluate the sentence importance. The algorithm is based on the intuition that the keyphrases represent the most important concepts of the text.

The proposed algorithm is different from directly scoring sentences through learning systems. These algorithms usually determine absolute importance for selecting a sentence, which is not true in all cases, since the importance is also determined by the allowed maximum limit of compression.

The algorithm extracts keyphrases contained in a text to be summarized. Based on the extracted keyphrases, each sentence within the text is ranked. The output summary is formed by extracting the sentences into the summary in order of decreasing ranks up to the specified summary length or percentage. In the proposed system, the importance of a sentence is determined by different heuristics. The current work introduces four scoring heuristics for achieving summarization goals, and shows the pros and cons for each.

Both of the word representation granularity level and its extracted morpho-syntactic features directly affect the performance of keyphrase extraction subsystem and hence the summarizer output. Section 4 reviews the lemma level Arabic analyzer. Next sections describe the two algorithm subsystems: Keyphrase Extractor and Sentence Extraction.

## 4. Arabic language Analyzer

Arabic is very rich in categorizing words, and hence, numerous techniques have been developed to evaluate the suitable representation level of Arabic words in IR systems. Two levels have been debated; root level and stem level. The main problem in selecting a root as a standard representation level in information retrieval systems is the over-semantic classification. Many words that do not have similar semantic interpretations are grouped into the same root. On the other hand, stem level suffers from under-semantic classification. Stem pattern may exclude many similar words sharing the same semantic properties. For example, Arabic broken plurals have stem patterns which differs from their singular patterns. In our work, we devoted towards lemma form representation level of Arabic words. Lemma refers to the set of all word forms that have the same meaning, and hence capture semantic similarities between words. On a word form conflation scale, lemma representation lies slightly above the (minimum) stem level, and below the (maximum) root level.

The keyphrase extractor used in our work is based on the existing Arabic keyphrase extractor AKE [6]. The original linguistic processing of AKE was based on an annotated Arabic corpus. To improve ambiguity resolution of Arabic POS tagging and increase the coverage scope of language analysis, we have replaced the corpus-based module with an accurate root-based lemmatizer module [5], which achieves maximum accuracy of 94.8%, and 89.15% for first seen documents. The input document is segmented into its constituent sentences based on the Arabic phrases delimiter characters such as comma, semicolon, colon, hyphen, and dot. Table (1) shows part of the output of the lemmatizer.

The accurate root-based Arabic lemmatizer performs the following tasks:
   a) Extracts POS tagging of the document words. Ambiguity is resolved using metadata about patterns, roots, and infixes' indications of Arabic words.
   b) Transforms inflected word form to its dictionary canonical lemma form. For nouns and adjectives, lemma form is the singular indefinite (masculine if possible) form, and for verbs, it is the perfective third person masculine singular form.
c) Extracts relevant morpho-syntactic features that support keyphrase extraction purposes.

Table 1: Sample of the Lemmatizer output

| Proposed Lemmatizer | | | | English | Arabic word |
|---|---|---|---|---|---|
| R | P | L | POS | | |
| عمد | تفتعل | اعتمد | VV | It (female) depends | تعتمد |
| | | معظم | particle | most | معظم |
| بلد | فعلان | بلد | NNS | countries | بلدان |
| علم | فاعل | عالم | DTNN | the world | العالم |
| | | الآن | RB | now | الآن |
| | | على | IN | on | على |
| خدم | استفعال | استخدام | NN | use | استخدام |
| نظم | افعل | نظام | DTNNS | the systems | الأنظمة |
| بني | مفعلة | مبني | DTJJ | based | المبنية |
| | | على | IN | on | على |
| حسب | فاعل | حاسب | DTNN | the computer | الحاسب |
| | | آلي | DTJJ | the automatic | الآلي |
| | | في | IN | in | في |
| نشأ | افعال | انشاء | NN | building | إنشاء |
| شغل | تفعيل | تشغيل | NN + | and operating | وتشغيل |
| | | صيانة | NN+ | and maintenance | وصيانة |
| شرع | مفاعيل | مشروع | NNS | projects | مشاريع |
| بني | فعلة | بنية | DTNN | the infra | البنية |
| سوس | افعل | اساس | DTJJ | the basic | الأساسية |
| خوص | فعلة | خاصة | DTJJ | the dedicated | الخاصة |
| | | بها | particle | for it | بها |
| | | في | IN | in | في |
| خلف | مفتعل | مختلف | NN | different | مختلف |
| قطع | فعال | قطاع | DTNNS | the sectors | القطاعات |
| مثل | | مثل | NN | like | مثل |
| قطع | فعال | قطاع | JJ | sectors | قطاعات |
| صنع | فعالة | صناعة | DTNN | the industry | الصناعة |
| زرع | فعالة | زراعة | DTNN + | and the agriculture | والزراعة |
| علم | تفعيل | تعليم | DTNN + | and the education | والتعليم |
| تجر | فعالة | تجارة | DFNN + | and the commerce | والتجارة |

## 5. Lemma based Keyphrase Extraction

The first step of the proposed summarizer algorithm is to extract indicative keyphrases of the document at a lemma level. We employ the existing Arabic keyphrase Extractor system [6]. The extractor is supplied with linguistic knowledge to enhance its efficiency instead of relying only on statistical information such as term frequency and distance.

The main modification done to AKE subsystem is the replacement of the annotated Arabic corpus with the accurate Arabic lemmatizer, to extract the required lexical features of the document words. The modifications to AKE are
a) Replacement of corpus-based analyzer with lemma-based analyzer.
b) Inclusion of Latin words and unrecognized words in keyphrases, they are treated as Arabic nouns.
c) Recognition of numerals.
d) Adding new sets of syntax rules that limits allowed word category sequences in candidate keyphrases.

The linguistic and statistical features are used to learn the Linear Discriminant Analysis classifier to extract relevant keyphrases. After modifications, the output of the keyphrase extractor subsystem is re-evaluated. The modified lemma-based Arabic keyphrase extractor is based on three main steps: Linguistic processing, candidate phrase extraction, and feature vector calculation. The following sections review these steps.

### 5.1 Candidate Phrases Extraction

The second step of the KP extractor is to extract all possible phrases of one, two, or three consecutive words that appear in a given document as n-gram terms. To extract effective candidate keyphrases, n-gram terms are then filtered according to syntactic rules that limits allowed POS sequences. Table (2) shows an example of the output 3-grams candidate keyphrases.

Table 2: Sample of extracted 3 grams keyphrases

| Sentence: "إن مشاريع التعليم عن بعد تعتبر من أهم تقنيات الاتصالات والمعلومات." | |
|---|---|
| Candidate Phrases (CP) | Abstract form of (CP) |
| مشاريع | مشروع |
| مشاريع التعليم | مشروع تعليم |
| التعليم | تعليم |
| التعليم عن بعد | تعليم عن بعد |
| بعد | بعد |
| تقنيات | تقنية |
| تقنيات الاتصالات | تقنية اتصال |
| تقنيات الاتصالات والمعلومات | تقنية اتصال معلومة |
| الاتصالات | اتصال |
| الاتصالات والمعلومات | اتصال معلومة |

### 5.2 Feature Vector Calculation

Each candidate phrase is then assigned a number of features used to evaluate its importance(.) The following features are adopted:
a) Normalized Phrase Words (NPW), which is the number of words in each phrase normalized to the maximum number of words in a phrase.
b) The Phrase Relative Frequency (PRF), which represents the frequency of abstract form of the

candidate phrase normalized by dividing it by the most frequent phrase in the given document.

c) The Word Relative Frequency (WRF): The frequency of the most frequent single abstract word in a candidate phrase (excluding article words), normalized by dividing it by the maximum number of repetitions of all phrase words in a given document.
d) Normalized Sentence Location (NSL), which measures the location of the sentence containing the candidate phrase within the document.
e) Normalized Phrase Location (NPL) feature is adopted to measure the location of the candidate phrase within its sentence.
f) Normalized Phrase Length (NPLen), which is the length of the candidate phrase (in words), divided by the number of words of its sentence.
g) Sentence Contain Verb (SCV). This feature has a value of zero if the sentence of the candidate phrase contains verb.
h) Is It Question (IIT): This feature has a value of one if the sentence of the candidate phrase is written in a question form.

In our work, we use the same LDA learning model of the corpus-based Arabic keyphrase extractor.

### 5.3 Evaluating lemma-based Arabic Keyphrase Extractor

Since we have changed the central language processing module of the keyphrase extractor, it was necessary to reevaluate its performance. The performance of the proposed lemma-based version of the Arabic KE is evaluated in two experiments. The first experiment uses same dataset described in [6] to compare the output keyphrases of the lemma-based version with those extracted by the corpus-based AKE, KP-Miner [4] (web link http://www.claes.sci.eg/coe_wm/kpminer), and Sakhr Keyword Extractor (web link http://www.sakhr.com/ Technology/ Keyword/ Default.aspx? sec=Technology &item= KeywordS). Table (3) shows extracted keyphrases from the four systems.

The results of the first experiment given in Table (4) show that the modified version of the keyphrase extractor has on average better performance than the corpus-based system in terms of precision and recall. The additional benefit we get is the increased language coverage of lemma-based system.

In the second experiment the data set is a parallel English/Arabic texts. Aligned texts from English UNICEF publication [23], and its corresponding Arabic translation [24], are used to compare extracted Arabic keyphrases using the proposed keyphrase extractor, to corresponding English extracted ones. Tables (5) and (6) show sample texts from both reports.

Table 3: Sample of results of the first experiment

| Lemma-based System | المرأة – **المرأة المصرية** ، مشاركة المرأة ، **نهضة المرأة** ، البحث العلمي، **المجلس القومي للمرأة** ، المطلوب ، **نهضة المرأة المصرية** ، **مشاركة المرأة المصرية** ، حقوق وواجبات |
|---|---|
| Corpus-based System | المرأة ، **المرأة المصرية** ، البحث العلمي _ **المجلس القومي للمرأة** ، المطلوب ، **نهضة المرأة المصرية** ، **مشاركة المرأة المصرية** ، حقوق وواجبات ، للمرأة المصرية وجود ، **مشاركات المرأة** |
| Kp-Miner | خريجات كليات – البحث العلمي – مجلسي الشعب – مجال القوي العاملة – مصر – الشوري – القومي – مصر – تعديل – إنجازات |
| Sakhr | **نهضة المرأة** ، مجلسي الشعب والشوري ، القوي العاملة ، وزارات الدولة – وزارة الصناعة – السلك الدبلوماسي – وزارة التضامن الاجتماعي – المؤسسات التشريعية والتنفيذية – طابعا دينيا – الأحزاب السياسية |

Table 4: Sample of average results of the first experiment

| # of Key phrases | Sakhr | | KP-Miner | | Original System | | Modified System | |
|---|---|---|---|---|---|---|---|---|
| | P | R | P | R | P | R | P | R |
| 10 | 0.19 | 0.12 | 0.34 | 0.20 | 0.65 | 0.40 | 0.71 | 0.52 |

Arabic texts are fed to lemma-based AKE, and KP-miner, while their corresponding English texts are fed to both of KEA [7] and Extractor [22] systems. Kea identifies candidate keyphrases by computing 4 feature values (TFxIDF), the first occurrence which is the percentage of the document preceding the first occurrence of the term, term length in words, and node degree of a candidate phrase which is the number of phrases in the candidate set that are semantically related to this phrase). Kea is available for download at
http://www.nzdl.org/Kea/index.html
Extractor is one of the major keyphrase extraction systems, with accuracy that ranges from 85% to 93% regardless of subject domain. Extarctor keyphrases, and summarization is available online at http://www.extractor.com/

Table 5 : English UNICEF sample document

| **UNICEF humanitarian action and resilience** |
|---|
| Guided by the Convention on the Rights of the Child, UNICEF in 2010 strengthened its core humanitarian policy to uphold the rights of children and women in crises. UNICEF reframed its Core Commitments for Children (CCCs) in Emergencies as the Core Commitments for Children in Humanitarian Action, reflecting wider shifts in UNICEF's own work in these contexts as well as the organization's commitment to humanitarian reform. Key changes include expanding the Core Commitments for Children to include preparedness before the onset of a crisis and adopting an early recovery approach during response – with disaster risk reduction integrated throughout. The Core Commitments for Children also moved from a focus on activities to broader strategic results that link |

humanitarian action to the fulfillment of children's and women's rights in each of UNICEF's programme sectors.
They also reflect the recognition that realizing these core commitments requires the contributions of a multitude of actors, including clusters. Thus reconceived, UNICEF's humanitarian action offers a potential platform for supporting resilience at the national and community levels. A few recent examples illustrate how this has manifested in emergency-affected countries.
The revised Core Commitments for Children also tighten the link between humanitarian action and development.
This stronger integration contributes to UNICEF's institutional flexibility – the nimbleness with which our programmes adjust to evolving situations. In addition, the sharpened focus on disaster risk reduction and local capacity development as explicit strategies contribute to communities' own flexibility in the face of multiple shocks, throughout the broader cycle of prevention, response and recovery. In Ethiopia, UNICEF has sup ported disaster risk reduction through a government-led, decentralized health extension programme to provide essential health and nutrition services. This programme has had a significant impact in the communities: Results show an increase in national treatment capacity of severe acute malnutrition from 135,000 cases per month in 2009 to 200,000 cases per month in 2010. Through the treatment of children suffering from malnutrition, those with severe acute malnutrition can now be identified earlier and receive life-saving treatment closer to home, thus helping reduce children's vulnerability.

Table 6 : Arabic UNICEF: A translation to English sample document in Table 5

| العمل الإنساني لليونيسف والصمود |
|---|
| واسترشاداً باتفاقية حقوق الطفل قامت اليونيسف في عام ٢٠١٠ بتعزيز سياستها الإنسانية الأساسية لدعم حقوق الأطفال والنساء في الأزمات. كما قامت بإعادة تأطير التزاماتها الأساسية نحو الأطفال في حالات الطوارئ والالتزامات الأساسية نحو الأطفال في العمل الإنساني، والتي تعكس تحولات أوسع في عمل اليونيسف في هذه السياقات، فضلاً عن التزام المنظمة بإجراء إصلاحات في المجال الإنساني. وتشمل التغييرات الرئيسية توسيع نطاق الالتزامات الأساسية نحو الأطفال لتشمل التأهب قبل بداية الأزمة واعتماد منهج الانتعاش المبكر خلال استجابة – مع دمج الحد من مخاطر الكوارث في جميع المراحل. كما تحول تركيز الالتزامات الأساسية نحو الأطفال من الأنشطة إلى الاستراتيجية الأوسع نطاقاً التي تربط العمل الإنساني بتلبية حقوق الطفل والمرأة في كل قطاعات برنامج اليونيسف. وهي تعكس أيضاً إدراك أن تحقيق هذه الالتزامات الأساسية يتطلب مساهمات من عدد كبير من الجهات الفاعلة، بما في ذلك مجموعات العمل الإنساني. ومن خلال إعادة فهمه، يقدم العمل الإنساني لليونيسف مدخلاً محتملا لدعم الصمود على المستويين الوطني والمجتمعي. وتوضح عدة أمثلة حديثة كيف يتجلى هذا في البلدان المتضررة من حالات الطوارئ. <br><br>وكذلك قامت الالتزامات الأساسية المنقحة نحو الأطفال بتعزيز العلاقة بين العمل الإنساني والتنمية. ويساهم هذا التكامل الوثيق في مرونة اليونيسف المؤسسية – وسهولة تكيف البرامج مع الأوضاع المستجدة. وبالإضافة إلى ذلك، فإن زيادة التركيز على الحد من مخاطر الكوارث وتطوير القدرات المحلية باعتبارها استراتيجيات واضحة تسهم في مرونة المجتمعات المحلية في مواجهة صدمات متعددة، في إطار الدوائر الأوسع للوقاية والانتعاش. وفي إثيوبيا، دعمت اليونيسف الحد من مخاطر الكوارث من خلال برنامج توفير الرعاية الصحية الذي تقوده الحكومة والغير مركزي لتقديم خدمات صحية وتغذوية أساسية. وقد كان لهذا البرنامج تأثير كبير على المجتمعات المحلية: وتظهر النتائج زيادة القدرات الوطنية على علاج سوء التغذية الحاد والشديد من ١٣٥٠٠٠ حالة في الشهر في عام ٢٠٠٩ إلى ٢٠٠٠٠٠ حالة في الشهر في عام ٢٠١٠ . ومن خلال معالجة الأطفال الذين يعانون من سوء التغذية، يمكن تحديد هؤلاء الذين يعانون من سوء التغذية الحاد الشديد مبكراً وتلقي العلاج المنقذ للحياة في أماكن قريبة من ديارهم، مما يساعد على الحد من تضرر الأطفال وتعرضهم للمخاطر. |

The results of the four systems shown in Table (7) reflect the complexity of evaluating keyphrase systems. For the two robust systems, only 25% of keyphrases extracted by KEA are similar to those extracted by Extractor. Our modified AKE has 57% similar keyphrases with Extractor, 58% similar Keyphrases with KP-Miner, and 33% similarity with KEA.

Table 7 : Keyphrases output of sample documents shown in Table 5 and Table 6

| EXTRACTOR | KEA | KP-Miner | Lemma AKE |
|---|---|---|---|
| core commitments | UNICEF | الالتزامات الأساسية | الأطفال |
| humanitarian action | Core Commitments | العمل الإنساني | اليونيسف |
| UNICEF | humanitarian action | مخاطر الكوارث | العمل |
| programme | Core Commitments for Children | اليونيسف | العمل الإنساني |
| rights | Commitments for Children | العمل الإنساني لليونيسف | الالتزامات |
| severe acute malnutrition | humanitarian | حالات الطوارئ | الالتزامات الأساسية |
| treatment | disaster risk reduction | الحد | حالات |
|  | disaster risk | المجتمعات المحلية | الحد |
|  | risk reduction | حقوق | مخاطر الكوارث |
|  | link humanitarian | الأطفال | برنامج |
|  | link humanitarian action | قامت | المجتمعات |
|  | severe acute | الصمود | العمل الإنساني لليونيسف |

## 6. Sentence Extraction

The sentence extraction algorithm exploits keyphrases that are automatically extracted from document text as the primary attributes of a sentence. Sentence ranking is determined by assigning scores to each sentence of the document based on extracted keyphrases. Different scoring schemes are adopted to achieve one or more goals of summarization. The output summary is then formed by extracting the sentences into the summary in order of decreasing ranks up to the specified summary length or percentage. In the proposed summarizer, extracted sentences are sequentially presented in the same sequence as the original text to preserve the information flow.
The following subsections describe four different heuristics for scoring sentences based on keyphrases. For each document, the top twelve extracted keyphrase are employed through the evaluation experiments. In all experiments, a compression ratio is set to 25% .

## 6.1 Summing keyphrases heuristic

Once extracted keyphrases scores have been computed, the next step of the algorithm is to score sentences. In the first heuristic, scoring is based on summing scores for all lemma keyphrase forms that appear in the sentence. The summing heuristic is given by

```
For each sentence S
    SUM all the scores of lemma form keyphrases
        contained in S.
Find NSS by Normalizing Sentences Scores by the maximum score .
```

Sentences are then ranked based on their Normalized Summing Scores (NSS), and an (n) percentage of sentences are extracted into a summary. The algorithm is greedy since it favors a sentence that contains many important concepts. For small percentage summarization, the algorithm succeeds to capture minimum number of sentences that carry most important topics of the document. Table (8) shows the output summary of the sample document.

Table 8 : Summary output based on NSS

| واسترشاداً باتفاقية حقوق الطفل قامت اليونيسف في عام 2010 بتعزيز سياساتها الإنسانية الأساسية لدعم حقوق الأطفال والنساء في الأزمات، كما قامت بإعادة تأطير التزاماتها الأساسية نحو الأطفال في حالات الطوارئ والالتزامات الأساسية نحو الأطفال في العمل الإنساني، كما تحول تركيز الالتزامات الأساسية نحو الأطفال من الأنشطة إلى الاستراتيجية الأوسع نطاقاً التي تربط العمل الإنساني بتلبية حقوق الطفل والمرأة في كل قطاعات برنامج اليونيسف. ومن خلال إعادة فهمه يقدم العمل الإنساني لليونيسف مدخلاً محتملا لدعم الصمود على المستويين الوطني والمجتمعي. وكذلك قامت الالتزامات الأساسية المنقحة نحو الأطفال بتعزيز العلاقة بين العمل الإنساني والتنمية. |
|---|

The main problems with summing based algorithm are its limited coverage and redundant sentence selections. A document usually consists of several topics. Main topics are described by many sentences, and hence have high scores. Supplementing concepts are usually described by few sentences, and hence have small scores. In summing based algorithm, many sentences that describe same (focus) topic dominate the selection. All sentences containing the main topic will get higher scores, which may lead to redundant sentences selection, and does not provide a room for other topics to appear in the summary.
We therefore defined a second heuristic based on counting keyphrases instead of summing their scores.

## 6.2 Counting keyphrases heuristic

One solution to cover the major topics of the document as much as possible, is to score sentences according to count of keyphrases instead of their scores. The counting heuristic is given by

```
For each sentence S
    COUNT all lemma form keyphrases contained in S
Find NCS by Normalizing the sentence Count scores by the maximum score .
```

This gives equal importance to all keyphrases, and hence concepts. Sentences that have more keyphrases are extracted and put in a summary. Table (9) shows summarization result for counting score heuristic.

Table 9 : Summary output based on NCS

| كما قامت بإعادة تأطير التزاماتها الأساسية نحو الأطفال في حالات الطوارئ والالتزامات الأساسية نحو الأطفال في العمل الإنساني، مع دمج الحد من مخاطر الكوارث في جميع المراحل. كما تحول تركيز الالتزامات الأساسية نحو الأطفال من الأنشطة إلى الاستراتيجية الأوسع نطاقاً التي تربط العمل الإنساني بتلبية حقوق الطفل والمرأة في كل قطاعات برنامج اليونيسف. ومن خلال إعادة فهمه يقدم العمل الإنساني لليونيسف مدخلاً محتملا لدعم الصمود على المستويين الوطني والمجتمعي. وكذلك قامت الالتزامات الأساسية المنقحة نحو الأطفال بتعزيز العلاقة بين العمل الإنساني والتنمية. . |
|---|

It is noticed that the algorithm tends to select longer sentences because they are more likely to achieve high count scores. This may be useful for a summary, where longer sentences tend to be more easily interpreted without surrounding context.
Count heuristic solves the problem of 'main topic' domination of the output summary. However, it does not guarantee the avoidance of redundant topic selection. The same problem exists also with 'summing heuristic'. Some authors repeat important sentences in many parts of the document with little word variations. The problem was solved in previous work [20] by adding a filter which removes redundant sentences based on cosine similarity measurement between all extracted sentences. In the current research, we have another approach discussed in the third heuristic.

## 6.3 Keyphrase coverage oriented heuristic

Both of the previously described two heuristics are based on scoring sentences, and don't guarantee complete coverage of all concepts of the document for a p% summary length. Also, very similar sentences can be extracted. The Coverage heuristic is given by

```
For each lemma form of keyphrase  K
  Increment the score of the first sentence containing K
Find NKS by Normalizing the sentence Key Scores by the maximum score.
```

In the coverage oriented algorithm, only one sentence is extracted for each keyphrase. It starts by high score keyphrases, extract the sentence that contains the first appearance of this keyphrase to summary, if it is not already exist. In this heuristic, only one sentence at most, is extracted for each keyphrase. The algorithm covers all the major topics of the document, and at the same time keeping redundancy to a minimum. Table 10 shows the resultant summary.

Table 10 Summary output based on NKS

| واسترشاداً باتفاقية حقوق الطفل قامت اليونيسف في عام ٢٠١٠ بتعزيز سياستها الإنسانية الأساسية لدعم حقوق الأطفال والنساء في الأزمات، كما قامت بإعادة تأطير التزاماتها الأساسية نحو الأطفال في حالات الطوارئ والالتزامات الأساسية نحو الأطفال في العمل الإنساني، مع دمج الحد من مخاطر الكوارث في جميع المراحل. ومن خلال إعادة فهمه يقدم العمل الإنساني لليونيسف مدخلاً محتملا لدعم الصمود على المستويين الوطني والمجتمعي. وفي إثيوبيا دعمت اليونيسف الحد من مخاطر الكوارث من خلال برنامج توفير الرعاية الصحية الذي تقوده الحكومة والغير مركزي لتقديم خدمات صحية وتغذوية أساسية. |
|---|

### 6.4 Merging heuristics

The fourth heuristic merges different scoring techniques to achieve a balance between summarization goals. This should lead to further improvement of the source text abstraction. The scoring of merging heuristic is formed by summing previous normalized scores, This is given by:

Merging Score = NSS + NCS + NKS

Evaluating automatic text summarization systems is not a straightforward process since it is an elusive property [10]. Since there are no Arabic standard summarization documents, we compare extracted Arabic sentences with existing well-established Extractor system. The following procedures are adopted through the experiment:
1- Apply the proposed system to generate a summary for Arabic texts, with compression ratio of 25% of the document sentences.
2- Extract the corresponding English sentences to have the "Equivalent English Summary".
3- Generate the "English Summary" of English texts corresponding to Arabic ones using Extractor system.
4- Compare the similarity of the Equivalent English Summary and English Summary.

Table (11) shows the 25% compression of a sample document Merging Heuristic Summaries. Table (12) presents its Equivalent English Summary. Extractor Summarization is given in Table (13). The average results of similarity between Extractor and the proposed system is nearly 66% for 25% compression summaries. More evaluation is still required to measure the similarity at different compression ratios.

Table 11: Summary output based on Merging Heuristic

| واسترشاداً باتفاقية حقوق الطفل قامت اليونيسف في عام ٢٠١٠ بتعزيز سياستها الإنسانية الأساسية لدعم حقوق الأطفال والنساء في الأزمات، كما قامت بإعادة تأطير التزاماتها الأساسية نحو الأطفال في حالات الطوارئ والالتزامات الأساسية نحو الأطفال في العمل الإنساني، مع دمج الحد من مخاطر الكوارث في جميع المراحل. كما تحول تركيز الالتزامات الأساسية نحو الأطفال من الأنشطة إلى الاستراتيجية الأوسع نطاقاً التي تربط العمل الإنساني بتلبية حقوق الطفل والمرأة في كل قطاعات برنامج اليونيسف. ومن خلال إعادة فهمه يقدم العمل الإنساني لليونيسف مدخلاً محتملا لدعم الصمود على المستويين الوطني والمجتمعي. |
|---|

Table 12: Equivalent English Summary with shaded area represent similarities with Extractor

| Guided by the Convention on the Rights of the Child, UNICEF in 2010 strengthened its core humanitarian policy to uphold the rights of children and women in crises. UNICEF reframed its Core Commitments for Children (CCCs) in Emergencies as the Core Commitments for Children in Humanitarian Action, with disaster risk reduction integrated throughout. The Core Commitments for Children also moved from a focus on activities to broader strategic results that link humanitarian action to the fulfillment of children's and women's rights in each of UNICEF's programme sectors. Thus reconceived, UNICEF's humanitarian action offers a potential platform for supporting resilience at the national and community levels. |
|---|

Table 13 : Summarization output of Extractor

| Guided by the Convention on the Rights of the Child, UNICEF in 2010 strengthened its core humanitarian policy to uphold the rights of children and women in crises. UNICEF reframed its Core Commitments for Children (CCCs) in Emergencies as the Core Commitments for Children in Humanitarian Action, reflecting wider shifts in UNICEF's own work in these contexts as well as the organization's commitment to humanitarian reform. The Core Commitments for Children also moved from a focus on activities to broader strategic results that link humanitarian action to the fulfillment of children's and women's rights in each of UNICEF's programme sectors. This programme has had a significant impact in the communities: Results show an increase in national treatment capacity of severe acute malnutrition from 135,000 cases per month in 2009 to 200,000 cases per month in 2010. |
|---|

## 7. Conclusions

In this research we have presented an Arabic summarization algorithm for extracting relevant sentences from free texts. The system exploits statistical and linguistic features to identify important keyphrases. Through experiments we show that different keyphrase based scoring schemes can direct the proposed sentence extractor towards one or more summarization goals.

**Tarek El-Shishtawy** is a Professor assistant at Faculty of Engineering, Benha University, Egypt. He participated in many Arabic computational Linguistic projects. Large Scale Arabic annotated Corpus, 1995, was one of important projects for Egyptian Computer Society, and Academy of Scientific Research and Technology, He has many publications in Arabic Corpus, machine translation, Text, and data Mining.

**Fatma El-Ghannam** is a researcher assistance at Electronics Research Institute – Cairo, Egypt. She has great research interests in Arabic language generation and analysis. Currently, she's preparing for a Ph.D. degree in NLP.